\RequirePackage{fix-cm}
\documentclass[smallextended]{svjour3}       
\smartqed  
\usepackage{graphicx}
\usepackage{url}
\usepackage{multirow}
\usepackage{xcolor}
\usepackage{subfig}
\usepackage{amsmath}
\usepackage{booktabs}
\begin{document}

\title{A Pragmatic Guide to Geoparsing Evaluation
\thanks{We gratefully acknowledge the funding support of the Natural Environment Research Council (NERC) PhD Studentship (Milan Gritta NE/M009009/1), EPSRC (Nigel Collier EP/M005089/1) and MRC (Mohammad Taher Pilehvar MR/M025160/1 for PheneBank). We also acknowledge Cambridge University linguists Mina Frost and Qianchu (Flora) Liu for providing expertise and verification (IAA) during dataset construction/annotation.}
}
%

\subtitle{Toponyms, Named Entity Recognition and Pragmatics}


\author{Milan Gritta         \and
        Mohammad Taher Pilehvar \and
        Nigel Collier
}


\institute{Milan Gritta \at
              \email{mg711@cam.ac.uk} 
            \and 
              Mohammad Taher Pilehvar \at
              \email{mp792@cam.ac.uk}
              \and 
              Nigel Collier \at
              \email{nhc30@cam.ac.uk}\\ \\
              Language Technology Lab (LTL)\\
              Department of Theoretical and Applied Linguistics (DTAL)\\
              University of Cambridge,  9 West Road, Cambridge CB3 9DP \\
}

\date{Received: date / Accepted: date}

 \maketitle

\begin{abstract}
Empirical methods in geoparsing have thus far lacked a standard evaluation framework describing the task, metrics and data used to compare state-of-the-art systems. Evaluation is further made inconsistent, even unrepresentative of real world usage by the lack of distinction between the \textit{different types of toponyms}, which necessitates new guidelines, a consolidation of metrics and a detailed toponym taxonomy with implications for Named Entity Recognition (NER) and beyond. To address these deficiencies, our manuscript introduces a new framework in three parts. Part 1) Task Definition: clarified via corpus linguistic analysis proposing a fine-grained \textit{Pragmatic Taxonomy of Toponyms}. Part 2) Metrics: discussed and reviewed for a rigorous evaluation including recommendations for NER/Geoparsing practitioners. Part 3) Evaluation Data: shared via a new dataset called \textit{GeoWebNews} to provide test/train examples and enable immediate use of our contributions. In addition to fine-grained Geotagging and Toponym Resolution (Geocoding), this dataset is also suitable for prototyping and evaluating machine learning NLP models. 
\end{abstract}

\section{Introduction}
Geoparsing aims to translate toponyms in free text into geographic coordinates. Toponyms are weakly defined as ``place names", however, we will clarify and extend this underspecified definition in Section \ref{taxonomy}. Illustrating with an example headline, ``\textit{Springfield robber escapes from Waldo County Jail. Maine police have launched an investigation.}", the geoparsing pipeline is (1) Toponym extraction [\textit{Springfield, Waldo County Jail, Maine}], this step is called \textit{Geotagging} and is a special case of NER; and (2) Disambiguating and linking toponyms to geographic coordinates [(45.39, -68.13), (44.42, -69.01), (45.50, -69.24)], this step is called \textit{Toponym Resolution} (also \textit{Geocoding}). Geoparsing is an essential constituent of many Geographic Information Retrieval (GIR), Extraction (GIE) and Analysis (GIA) tasks such as determining a document's geographic scope \cite{steinberger2013introduction}, Twitter-based disaster response \cite{de2018taggs} and mapping \cite{avvenuti2018crismap}, spatio-temporal analysis of tropical research literature \cite{palmblad2017spatiotemporal}, business news analysis \cite{abdelkader2015brands}, disease detection and monitoring \cite{allen2017global} as well as analysis of historical events such as the Irish potato famine \cite{tateosian2017tracking}. Geoparsing can be evaluated in a highly rigorous manner, enabling a robust comparison of state-of-the-art (SOTA) methods. This manuscript provides the end-to-end \textit{Pragmatic Guide to Geoparsing Evaluation} for that purpose. End-to-end means to (1) critically review and extend the definition of toponyms, i.e. \textit{what} is to be evaluated and \textit{why} it is important; (2) review, recommend and create high-quality open resources to expedite research; and (3) outline, review and consolidate metrics for each stage of the geoparsing pipeline, i.e. \textit{how} to evaluate. \\

Due to the essential NER component in geoparsing systems \cite{santos2015using,delozier2015gazetteer,karimzadeh2013geotxt,gritta2017s,jurgens2015geolocation}, our investigation and proposals have a strong focus on NER's \textit{location extraction} capability. We demonstrate that off-the-shelf NER taggers are inadequate for location extraction due to the lack of ability to extract and classify the pragmatic types of toponyms (Table \ref{summary}). In an attempt to assign coordinates to an example sentence, ``\textit{A French bulldog bit an Australian tourist in a Spanish resort.}", current NER tools fail to differentiate between the literal and associative uses of these adjectival toponyms\footnote{Throughout the paper, we use the term \textit{Literal} to denote a toponym and/or its context that refers directly to the \textit{physical} location and the term \textit{Associative} for a toponym and/or its context that is only \textit{associated} with a place. Full details in Section \ref{taxonomy}.}. A more detailed example analysed in Table \ref{comparetable} and a survey of previous work in Section \ref{related} show that the definition and handling of toponyms is inconsistent and unfit for advanced geographic NLP research. In fact, beyond a limited ``{place name}" definition, a deep pragmatic/contextual toponym semantics has not yet been defined in Information Extraction, to our best knowledge. This underspecification results in erroneous and unrepresentative real-world extraction/classification of toponyms incurring both \textit{precision errors} and \textit{recall errors}. To that end, we propose a \textit{Pragmatic Taxonomy of Toponyms} required for a rigorous geoparsing evaluation, which includes the recommended datasets and metrics. 

\paragraph{Why a Pragmatic Guide} Pragmatics \cite{pustejovsky1991generative} is the linguistic theory of generative approach to word meaning, i.e. how context contributes to and changes the semantics of words and phrases. This is the first time, to our best knowledge, that the definition of fine-grained toponym types has been quantified in such detail using a representative sample of general topic, globally distributed news articles. We also release a new \textit{GeoWebNews} dataset to challenge researchers to develop Machine Learning (ML) algorithms to evaluate classification/tagging performance based on deep pragmatics rather than shallow syntactic features. Section \ref{back} gives a background on Geoparsing, NER, GIE and GIR. We present the new taxonomy in Section \ref{taxonomy}, describing and categorising toponym types. In Section \ref{sef}, we conduct a comprehensive review of current evaluation methods and justify the recommended framework. Finally, Section \ref{geowebnews} introduces the GeoWebNews dataset, annotation and resources. We also evaluate geotagging and toponym resolution on the new dataset, illustrating the performance of several sequence tagging models such as SpacyNLP and Google NLP.

\subsection{Summary of the most salient findings}
Toponym semantics have been underspecified in NLP literature.
Toponyms can refer to physical places as well as entities associated with a place as we outline in our proposed taxonomy. Their distribution in a sample of 200 news articles is 53\% literal and 47\% associative. Until now, this type of fine-grained toponym analysis was not conducted. We provide a dataset annotated by linguists (including computational) enabling immediate evaluation of our proposals. GeoWebNews.xml can be used to evaluate Geotagging, NER, Toponym Resolution and to develop ML models from limited training data. A total of 2,720 toponyms were annotated with Geonames\footnote{\url{https://www.geonames.org/}}. Data augmentation was evaluated with an extra 3,460 annotations although effective implementation remains challenging. We also found that popular NER taggers appear not to use contextual information, relying instead on the entity's primary word sense (see Table \ref{comparetable}). We show that this issue can be addressed by training an effective geotagger from limited training data (F-Score=88.6), outperforming Google Cloud NLP (F-Score=83.2) and Spacy NLP (F-Score=74.9). In addition, effective 2-class (Literal versus Associative toponyms) geotagging is also feasible (F-Score=77.6). The best toponym resolution scores for GeoWebNews were 95\% accuracy@161km, AUC of 0.06 and a Mean Error of 188km. Finally, we provide a critical review of available metrics and important nuances of evaluation such as database choice, system scope, data domain/distribution, statistical testing, etc. All recommended resources are available on GitHub\footnote{\url{https://github.com/milangritta/Pragmatic-Guide-to-Geoparsing-Evaluation}}.

\section{Background}
\label{back}
Before we critically review \textit{how} to rigorously evaluate geoparsing and introduce a new dataset, we first need to clarify \textit{what} is to be evaluated and \textit{why}. We focus on the pragmatics of toponyms for fine-grained geoparsing of events described in text. This requires differentiating literal from associative types as well as increasing toponym recall by including entities ignored by current models. When a word spells like a place, i.e. shares its orthographic form, this does not mean it \textit{is} a place or has equivalent meaning, for example: ``\textit{Paris} (a person) said that \textit{Parisian} (associative toponym) artists don't have to live in \textit{Paris} (literal toponym)." and ``\textit{Iceland} (a UK supermarket) doesn't sell \textit{Icelandic} (associative toponym) food, it's not even the country of \textit{Iceland} (literal toponym)." In order to advance research in toponym extraction and other associated NLP tasks, we need to move away from the current practice of seemingly ignoring the context of a toponym, relying on the entity's dominant word sense and morphological features, treating toponyms as semantically equivalent. The consequences of this simplification are disagreements and incompatibilities in toponym evaluation leading to unrepresentative real-world performance. It is difficult to speculate about the reason for this underspecification, whether it is the lack of available quality training data leading to lower traction in the NLP community or the satisfaction with a simplified approach. However, we aim to encourage active research and discussions through our contributions.

\subsection{Geographic datasets and the pragmatics of toponyms}
Previous work in annotation of geographic NLP datasets constitutes our primary source of enquiry into recent research practices, especially the lack of linguistic definition of toponym types. An early specification of an Extended Named Entity Hierarchy \cite{sekine2002extended}  was based only on \textit{geographic feature types}\footnote{\url{https://nlp.cs.nyu.edu/ene/version7_1_0Beng.html}} i.e. address, country, region, water feature, etc. Geoparsing and NER require a deeper contextual perspective based on how toponyms are used in practice by journalists, writers or social media users, something a static database lookup cannot determine. CoNLL 2002 \cite{sang2002introduction} and 2003 \cite{tjong2003introduction} similarly offer no semantic definition of a toponym beyond what is naively thought of as a location, i.e. an entity spelled like a place and a location as its primary word sense. Schemes such as ACE \cite{doddington2004automatic} bypass toponym type distinction, classifying entities such as governments via a simplification to a single tag \textit{GPE: A Geo-Political Entity}. Modern NER parsers such as Spacy \cite{honnibal-johnson:2015:EMNLP} use similar schemes \cite{weischedel2013ontonotes} to collapse different taxonomic types into a single tag avoiding the need for a deeper understanding of context. A simplified tag set (LOC, ORG, PER, MISC) based on Wikipedia \cite{nothman2013learning} is used by NER taggers such as Illinois NER \cite{redman2016illinois} and Stanford NLP \cite{manning-EtAl:2014:P14-5}, featured in Table \ref{comparetable}. The table shows the limited classification indicating weak and inconsistent usage of context.\\

\label{related}
The SpatialML \cite{mani2010spatialml} scheme is focused on spatial reasoning e.g. \textit{X location north of Y}. Metonymy \cite{markert2002metonymy}, which is a substitution of a related entity for a concept originally meant, was acknowledged but not annotated due to the lack of training of Amazon Mechanical Turk annotators. Facilities were always tagged in the SpatialML corpus \textit{regardless of the context} in which they're being used. The corpus is available at a cost of \$500-\$1,000. The Message Understanding Conferences (MUC) \cite{hirschman1998evolution} have historically not tagged adjectival forms of locations such as \textit{``American} exporters". We assert that there is no difference between that and \textit{``U.S.} exporters", which would almost certainly be annotated. The Location Referring Expression corpus \cite{matsuda2015annotating} has annotated toponyms including locational expressions such as parks, buildings, bus stops and facilities in 10,000 Japanese tweets. Systematic polysemy \cite{alonso2013annotation} has been taken into account for \textit{facilities}, but not extended to other toponyms. GeoCLEF \cite{gey2005geoclef} (Geographic Cross Language Evaluation Forum) focused on Multilingual GIR evaluation. Geoparsing specifically, i.e. Information Extraction was not investigated. Toponym types were not linguistically differentiated despite the multi-year project's scale. This conclusion also applies to Spatial Information Retrieval and Geographical Ontologies \cite{jones2002spatial} (called SPIRIT) project, the focus of which was not the evaluation of Information Extraction or Toponym Semantics but classical GIR.\\

The WoTR corpus \cite{delozier2016data} of historical US documents also did not define toponyms. However, browsing the dataset, expressions such as ``Widow Harrow's house'' and ``British territory'' were annotated. In Section \ref{taxonomy}, we shall claim this is beyond the scope of toponyms, i.e. ``house" and ``territory" should not be tagged. The authors do acknowledge, but \textit{do not annotate} metonymy, demonyms and nested entities. Systematic polysemy such as metonymy should be differentiated during toponym extraction and classification, something acknowledged as a problem more than ten years ago \cite{leveling2008metonymy}. Section \ref{taxonomy} elaborates on the taxonomy of toponyms beyond metonymic cases. Geocorpora \cite{wallgrun2018geocorpora} is a Twitter-based geoparsing corpus with around 6,000 toponyms with buildings and facilities annotated. The authors acknowledge that toponyms are frequently used in a metonymic manner, however, these cases have not been annotated after browsing the open dataset. Adjectival toponyms have also been \textit{left out}. We show that these constitute around 13\% of all toponyms thus should be included to boost recall. \\

The LGL corpus \cite{lieberman2010geotagging} loosely defines toponyms as ``spatial data specified using text". The evaluation of an accompanying model focused on toponym resolution. Authors agree that standard Named Entity Recognition is inadequate for geographic NLP tasks. It is often the case that papers emphasise the geographic ambiguity of toponyms but not their semantic ambiguity. The CLUST dataset \cite{lieberman2011multifaceted} by the same author, describes toponyms simply as ``textual references to geographic locations". Homonyms are discussed as is the low recall and related issues of NER taggers, which makes them unsuitable for achieving high geotagging fidelity. Metonymy was not annotated, some adjectival toponyms have been tagged though sparsely and inconsistently. There is no distinction between literal and associative toponyms. Demonyms were tagged but with no special annotation hence treated as ordinary locations with no descriptive statistics offered. TR-News \cite{kamalloo2018coherent} is a quality geoparsing corpus despite the paucity of annotation details or IAA figures in the paper. A brief analysis of the open dataset showed that embedded toponyms, facilities and adjectival toponyms were annotated, which substantially increases recall, although no special tags were used hence unable to gather descriptive statistics. Homonyms, coercion, metonymy, demonyms and languages were not annotated and nor was the distinction between literal, mixed and associative toponyms. With that, we still recommended it as a suitable resource for geoparsing in the latter sections.
 
\paragraph{PhD Theses} are themselves comprehensive collections of a large body of relevant research and therefore important sources of prior work. Despite this not being the convention in NLP publishing, we outline the prominent PhD theses from the past 10+ years to show that toponym types have not been organised into a pragmatic taxonomy and that evaluation metrics in geocoding are in need of review and consolidation. We also cite their methods and contributions as additional background for discussions throughout the paper. The earliest comprehensive research on toponym resolution originated in (Leidner, 2008) \cite{leidner2008toponym}. Toponyms were specified as ``names of places as found in a text". The work recognised the ambiguity of toponyms in different contexts and was often cited by later research papers though until now, these linguistic regularities have not been formally and methodically studied, counted, organised and released as high fidelity open resources. A geographic mining thesis (Martins, 2008) \cite{da2008geographically} defined toponyms as ``geographic names" or ``place names". It mentions homonyms, which are handled with personal name exclusion lists rather than learned by contextual understanding. A Wikipedia GIR thesis (Overell, 2009) \cite{overell2009geographic} has no definition of toponyms and limits the analysis to \textit{nouns only}. The GIR thesis (Andogah, 2010) \cite{andogah2010geographically} discusses the geographic hierarchy of toponyms as found \textit{in gazetteers}, i.e. feature types instead of linguistic types. A toponym resolution thesis (Buscaldi, 2010) \cite{buscaldi2010toponym} describes toponyms as ``place names", once again mentions metonymy without handling these cases citing lack of resources, which our work provides.\\

The Twitter geolocation thesis (Han, 2014) \cite{han2014improving} provides no toponym taxonomy, nor does the Named Entity Linking thesis (Santos, 2013) \cite{dos2013linking}. A GIR thesis (Moncla, 2015) \cite{moncla2015automatic} defines a toponym as a spatial named entity, i.e. a location somewhere in the world bearing a proper name, discusses syntactical rules and typography of toponyms but not their semantics. The authors recognise this as an issue in geoparsing but no solution is proposed. The GIA thesis (Ferr{\'e}s, 2017) \cite{ferres2017knowledge} acknowledges but doesn't handle cases of metonymy, homonymy and non-literalness while describing a toponym as ``a geographical place name". Recent Masters theses also follow the same pattern such as a toponym resolution thesis (Kolkman, 2015) \cite{kolkman2015cross}, which says a toponym is a ``word of phrase that refers to a location". While none of these definitions are incorrect, they are very much underspecified. Another Toponym Resolution thesis (DeLozier, 2016) \cite{delozier2016data} acknowledges relevant linguistic phenomena such as metonymy and demonyms, however, no resources, annotation or taxonomy is given. Toponyms were established as ``named geographic entities". This background section presented a multitude of research contributions using, manipulating and referencing toponyms, however, without a deep dive into their pragmatics, i.e. what \textit{is} a toponym from a \textit{linguistic point of view} and the practical NLP implications of that. Without an agreement on the \textit{what}, \textit{why} and \textit{how} of geoparsing, the evaluation of SOTA systems cannot be consistent and robust.

\section{A Pragmatic Taxonomy of Toponyms}
\label{taxonomy}
\begin{figure*}[t]
\centering
\includegraphics[width=\textwidth]{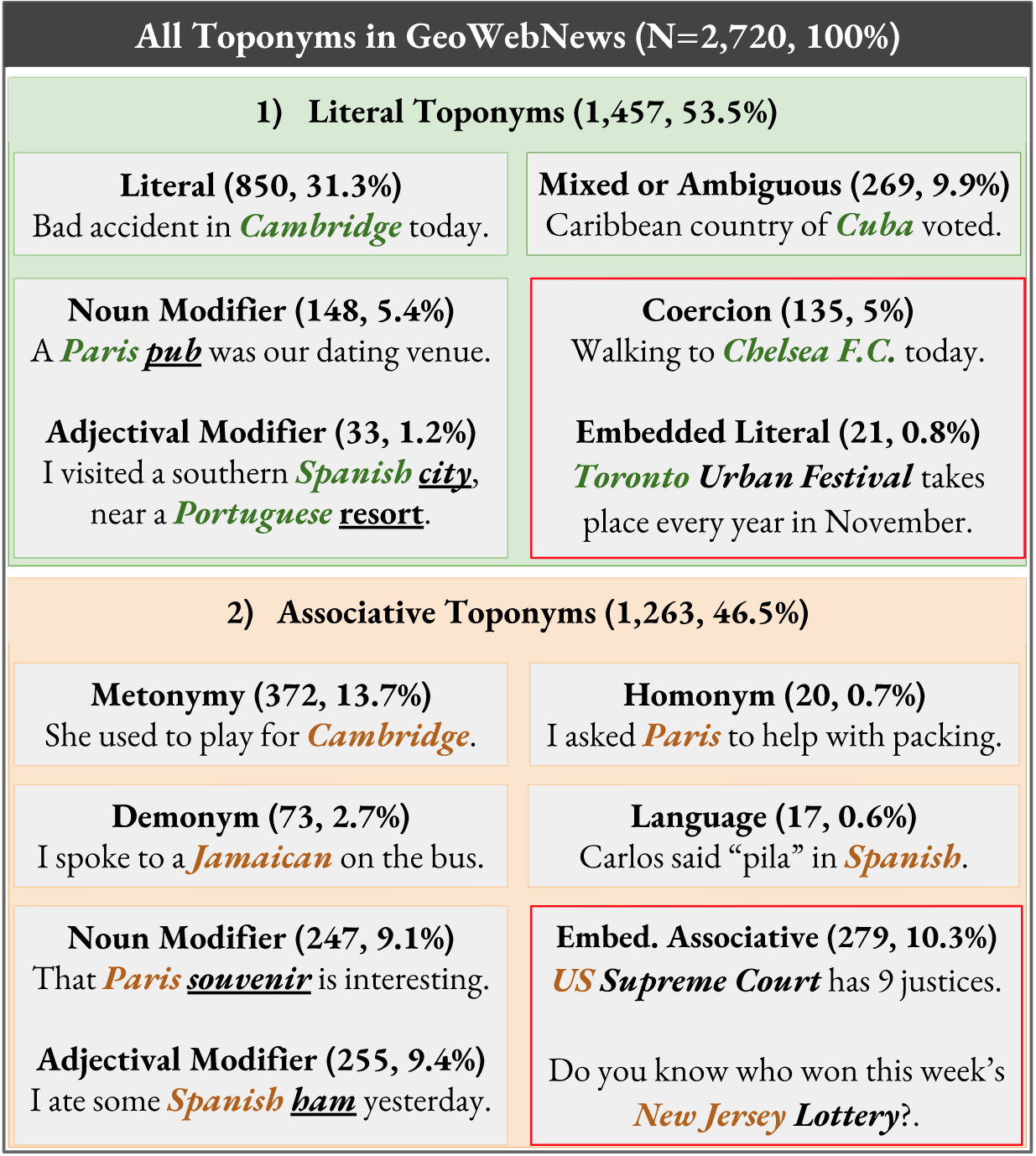}
\caption{The Pragmatic Taxonomy of Toponyms. A red border denotes \textit{Non-Toponyms}. Classification algorithm: If the context indicates a literal or is ambiguous/mixed, then the type is literal. If the context is associative, then (a) for non-modifiers the toponym is associative (b) for modifiers, if the head is mobile and/or abstract, then the toponym is associative, otherwise it is literal.}
\label{literal}
\end{figure*}

While the evaluation metrics, covered in Section \ref{sef}, are relevant only to geoparsing, Sections \ref{taxonomy} and \ref{geowebnews} have implications for Core NLP tasks such as NER. In order to introduce the Toponym Taxonomy, shown in Figure \ref{literal}, we start with a location. \textit{A location} is any of the potentially infinite physical points on Earth identifiable by \textit{coordinates}. With that in mind, \textit{a toponym is any named entity that labels a \textit{particular} location.} Toponyms are thus a subset of locations as most locations do not have proper names. Further to the definition and extending the work from the Background section, toponyms exhibit various degrees of \textit{literalness} as their \textit{referents} may not be physical locations but other entities as is the case with metonyms, languages, homonyms, demonyms, some embedded toponyms and associative modifiers. \\

Structurally, toponyms occur within clauses, which are the smallest grammatical units expressing a full proposition. Within clauses, which serve as the context, toponyms are embedded in noun phrases (NP). A toponym can occur as the \textit{head} of the NP, for example ``Accident in \textit{Melbourne}.'' Toponyms also frequently \textit{modify} NP heads. Modifiers can occur before or after the NP head such as in ``President of \textit{Mongolia}" versus ``\textit{Mongolian} President" and can have an \textit{adjectival} form ``\textit{European} cities" or a \textit{noun} form ``\textit{Europe}'s cities". In theory, though not always in practice, the classification of toponym types is driven by (1) \textit{the semantics of the NP}, which is conditional on (2) \textit{the NP context} of the surrounding clause. These types may be classified using a hybrid approach \cite{dong2015hybrid}, for example. It is this interplay of semantics and context, seen in Table \ref{summary}, that determines the type of the following toponyms (literals=\textbf{bold}, associative=\textit{italics}): ``The \textbf{Singapore} project is sponsored by \textit{Australia}." and ``He has shown that in \textbf{Europe} and last year in \textbf{Kentucky}." and ``The soldier was operating in \textbf{Manbij} with \textit{Turkish} troops when the bomb exploded." As a result of our corpus linguistic analysis, we propose \textit{two top-level} taxonomic types (a) \textit{literal}: where something is happening or is physically located; and (b) \textit{associative}: a concept that is associated with a toponym (Table \ref{summary}). We also assert that for applied NLP, it is sufficient and feasible to distinguish between literal and associative toponyms.

\subsection{Non-Toponyms}
\label{embedded} There is a group of entities that are currently not classified as toponyms, denoted as \textit{Non-Toponyms} in this paper. We shall assert, however, that these are in fact equivalent to ``regular" toponyms. We distinguish between three types: a) \textit{Embedded Literals} such as ``The \textit{British} Grand Prix" and \textit{``Louisiana} Purchase" b) \textit{Embedded Associative} toponyms, for example \textit{``Toronto} Police" and \textit{``Brighton} City Council'' and c) \textit{Coercion}, which is when a polysemous entity has its less dominant word sense \textit{coerced} to the \textit{location} class by the context. Failing to extract Non-Toponyms lowers real-world recall, missing out on valuable geographical data. In our diverse and broadly-sourced dataset, Non-Toponyms constituted a non-trivial 16\% of all toponyms.

\begin{table}[t]
\centering
\setlength{\tabcolsep}{7pt}
\begin{tabular}{l|c|c}
\toprule
\textbf{Toponym Type} & \textbf{NP Semantics Indicates} & \textbf{NP Context Indicates} \\
\midrule
{\color{teal}\textbf{Literals}} & \textbf{{\color{teal}Noun Literal Type}} & \textbf{{\color{teal}Literal Type}} \\
{\color{teal}\textbf{Literal Modifiers}} & \textbf{{\color{teal} Noun/Adjectival Literal}} & \textbf{{\color{teal}Literal} or {\color{orange}Associative}$^\dagger$} \\
{\color{teal}\textbf{Mixed}} & \textbf{{\color{teal} Noun/Adjectival Literal}} & \textbf{{\color{teal}Ambiguous} or {\color{teal}Mixed}} \\
{\color{teal}\textbf{Coercion}} & \textbf{{\color{red} Non-Toponym}} & \textbf{{\color{teal}Literal Type}} \\
{\color{teal}\textbf{Embedded Literal}} & \textbf{{\color{red}Non-Toponym}} & \textbf{{\color{teal}Literal Type}} \\ \midrule
{\color{orange}\textbf{Embedded NonLit}} & \textbf{{\color{red}Non-Toponym}} & \textbf{{\color{orange}Associative Type}} \\
{\color{orange}\textbf{Metonymy}} & \textbf{{\color{teal}Noun Literal Type}} & \textbf{{\color{orange}Associative Type}} \\ 
{\color{orange}\textbf{Languages}} & \textbf{{\color{teal}Adjectival Literal Type}} & \textbf{{\color{orange}Associative Type}} \\
{\color{orange}\textbf{Demonyms}} & \textbf{{\color{teal} Adjectival Literal Type}} & \textbf{{\color{orange}Associative Type}} \\
{\color{orange}\textbf{Non-Lit Modifiers}} & \textbf{{\color{teal} Noun/Adjectival Literal}} & \textbf{{\color{orange}Associative Type}} \\
{\color{orange}\textbf{Homonyms}} & \textbf{{\color{teal}Noun Literal Type}} & \textbf{{\color{orange}Associative Type}} \\\bottomrule
\end{tabular}
\caption{The interplay between context and semantics determines the type. The top five are the literals, the bottom six are the associative types. Examples of each type can be found in Figure \ref{literal}. $^\dagger$NP \textbf{head} must be strongly indicative of a literal type, e.g.: ``The \textit{British} \textbf{weather} doesn't seem to like us today."}
\label{summary}
\end{table}

\subsection{Literal Toponyms}
\label{lit_loc}
These types refer to places \textit{where something is happening or is physically located}. This subtle but important distinction from associative toponyms allows for higher quality geographic analysis. For instance, the phrase ``\textit{Swedish} people" (who could be anywhere) is \textit{not} the same as ``people \textit{in Sweden}" so we differentiate this group from the associative group. Only the latter mention refers to Swedish ``soil'' and can/should be processed separately.

\paragraph{A \bf Literal} is what is most commonly and too narrowly thought of as a location, e.g. ``Harvests in \textit{Australia} were very high." and ``\textit{South Africa} is baking in 40C degree heat." For these toponyms, the semantics and context both indicate it is a literal toponym, which refers directly to a physical location.

\paragraph{\bf Coercion}
\label{coercion} refers to polysemous entities typically classified as Non-Toponyms, which in a \textit{literal context} have their word sense coerced to (physical) \textit{location}. More formally, coercion is ``an observation of grammatical and semantic incongruity, in which a syntactic structure places requirements on the types of lexical items that may appear within it."\cite{ziegeler2007word} Examples include ``The \textit{University of Sussex, Sir Isaac Newton (pub), High Court} is our meeting place." and ``I'm walking to \textit{Chelsea F.C., Bell Labs, Burning Man.}" Extracting these toponyms increases recall and allows for a \textit{very precise location} as these toponyms tend to have a small geographic footprint.

\paragraph{\bf Mixed Toponyms} typically occur in an ambiguous context, e.g. ``\textit{United States} is generating a lot of pollution.'' or ``\textit{Sudan} is expecting a lot of rain.'' They can also simultaneously activate a literal \textit{and} an associative meaning, e.g. ``The north African country of \textit{Libya} announced the election date.'' These cases sit somewhere between literal and associative toponyms, however, we propose to include them in the literal group.

\paragraph{\bf Embedded Literals} are Non-Toponyms nested within larger entities such as \textit{``Toronto} Urban Festival", \textit{``London} Olympics", \textit{``Monaco} Grand Prix" and are often extracted using a 'greedy algorithm'. They are semantically, though not syntactically, equivalent to Literal Modifiers. If we ignored the case, the meaning of the phrase would not change, e.g. ``\textit{Toronto} urban festival".

\begin{figure}[t]
\centering
\includegraphics[width=\textwidth]{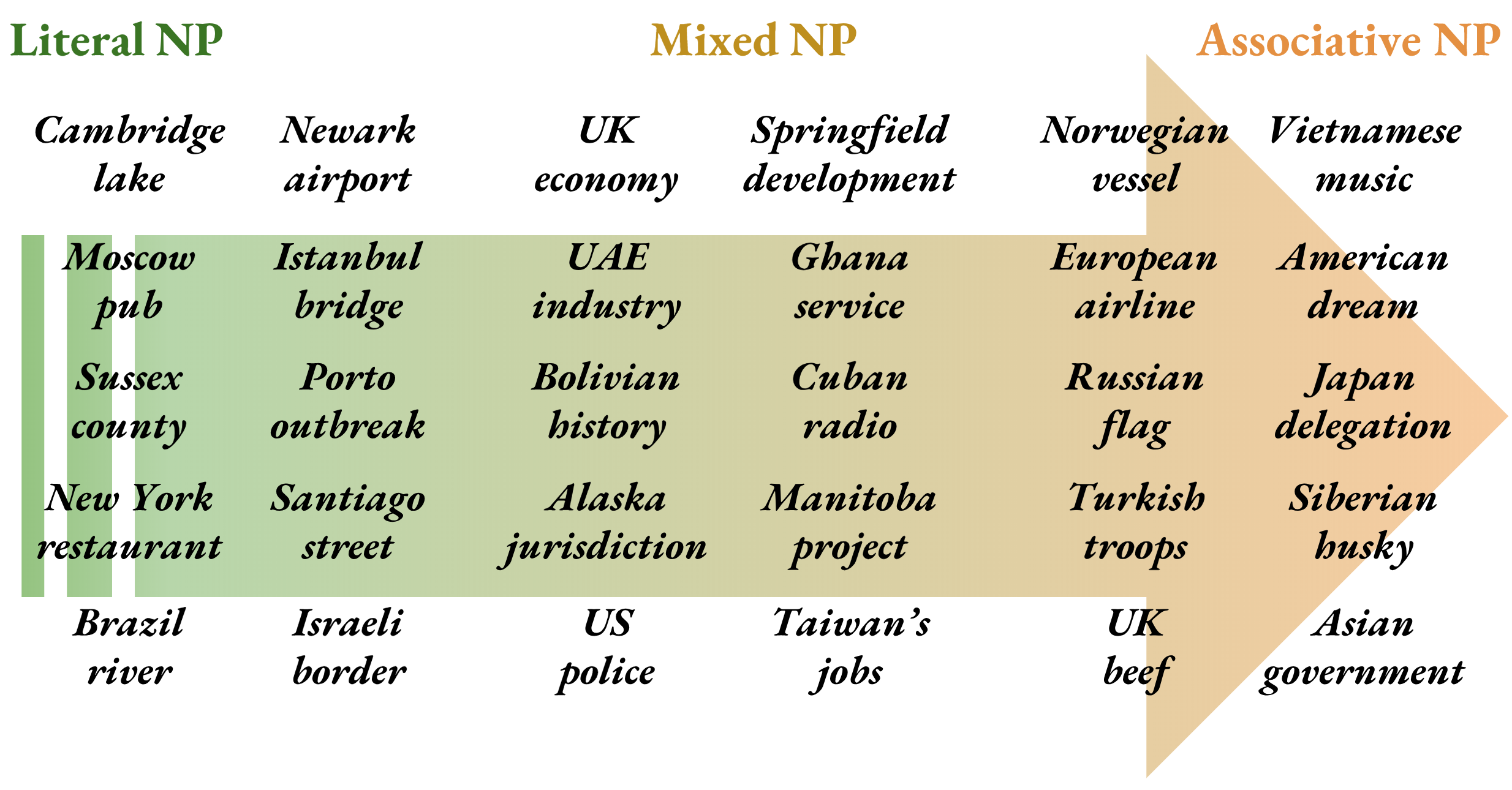}
\caption{Example noun phrases ranging from Literal to Mixed to Associative. The further to the right, the more 'detached' the \textit{NP referent} becomes from its physical location. Literal heads tend to be \textit{concrete} (elections, accidents) and \textit{static} (buildings, natural features) while associative heads are more \textit{abstract} (promises, partnerships) and \textit{mobile} (animals, products). In any case, \textit{context} is the main indicator of type and needs to be combined with \textit{NP semantics}.}
\label{heads}
\end{figure}

\paragraph{\bf Noun Modifiers}
\label{noun_mod} are toponyms that modify \textit{literal heads} (Figure \ref{heads}), e.g. ``You will find the UK [\textit{lake, statue, valley, base, airport}] there." and ``She was taken to the South Africa [\textit{hospital, border, police station}]". The context, however, needn't always be literal, for instance ``An \textit{Adelaide} court sentenced a murderer to 25 years.'' or ``The \textit{Vietnam} office hired 5 extra staff.'' providing the head is literal. Noun modifiers can also be placed after the head, for instance ``We have heard much about the stunning caves of \textit{Croatia}.'' 

\paragraph{\bf Adjectival Modifiers}
\label{adj_mod} exhibit much the same pattern as noun modifiers except for the \textit{adjectival form} of the toponym, for example, ``It's freezing in the \textit{Russian} tundra.", ``\textit{British} ports have doubled exports.'' or ``\textit{American} schools are asking for more funding." Adjectival modifiers are frequently and incorrectly tagged as \textit{nationalities or religious/political groups}\footnote{\url{https://spacy.io/usage/} and \url{http://corenlp.run/}} and sometimes ignored\footnote{\url{http://services.gate.ac.uk/annie/} and IBM NLP Cloud in Table \ref{comparetable}.} altogether. Approximately 1 out of 10 adjectival modifiers is literal.

\subsection{Associative Toponyms}
\label{associative}
Toponyms frequently refer to or are used to modify \textit{non-locational concepts} (NP heads), which are \textit{associated} with locations rather than directly referring to their physical presence. This can occur by substituting a non-locational concept with a toponym (metonymy) or via a demonym, homonym or a language reference. Some of these instances look superficially like modifiers leading to frequent NER errors.

\paragraph{\bf Demonyms}
\label{demonyms} \cite{roberts2011germans} are derived from toponyms and denote the inhabitants of a country, region or city. These persons are \textit{associated} with a location and have been on occasion, sparsely rather than exhaustively, annotated \cite{lieberman2010geotagging}. Examples include ``I think he's \textit{Indian}.'', which is equivalent to ``I think he's an \textit{Indian} citizen/person.'' or ``An \textit{American} and a \textit{Briton} walk into a bar ..."

\paragraph{\bf Languages}
\label{languages} can sometimes be confused for adjectival toponyms, e.g. ``How do you say pragmatics in \textit{French, Spanish, English, Japanese, Chinese, Polish?}'' Occurrences of languages should not be interpreted as modifiers, another NER error stemming from a lack of contextual understanding. This is another case of a concept associated with a location that should not require coordinates.

\paragraph{\bf Metonymy}
\label{metonymy} is a figure of speech whereby a concept that was originally intended gets \textit{substituted} with a \textit{related} concept, for example ``\textit{Madrid} plays \textit{Kiev} today.", substituting sports teams with toponyms. Similarly, in ``\textit{Mexico} changed the law.'', the likely latent entity is the \textit{Mexican government}. Metonymy was previously found to be a frequent phenomenon, around 15-20\% of place mentions are metonymic \cite{markert2007semeval,gritta2017vancouver,leveling2008metonymy}. In our dataset, it was 13.7\%.

\paragraph{\bf Noun Modifiers}
are toponyms that modify associative noun phrase heads in an associative context, for instance ``\textit{China} exports slowed by 7 percent.'' or ``\textit{Kenya's} athletes win double gold.'' Noun modifiers also occur after the head as in ``The President \textit{of Armenia} visited the Embassy of \textit{the Republic of Armenia} to the \textit{Vatican}.". Note that the event did \textit{not} take place in Armenia but the Vatican, potentially identifying the wrong event location.

\paragraph{\bf Adjectival Modifiers}
are sporadically covered by NER taggers (Table \ref{comparetable}) or tagging schemes \cite{hirschman1998evolution}. They are semantically identical to associative noun modifiers except for their adjectival form, e.g. ``\textit{Spanish} sausages sales top €2M.'', ``We're supporting the \textit{Catalan} club." and ``\textit{British} voters undecided ahead of the Brexit referendum." 

\begin{table*}[t]
\centering
\setlength{\tabcolsep}{1.3pt}
\begin{tabular}{l|c|cccccc}
\toprule
\textbf{TOPONYM (TYPE)} & \textbf{LABEL} & \textbf{GOOG}. & \textbf{SPACY} & \textbf{STANF}. & \textbf{ANNIE} & \textbf{ILLIN}. & \textbf{IBM}\\
\midrule
\textbf{Milan} (Homomymy) & \textbf{Assoc.} & \textbf{\underline{Literal}} & \textbf{\underline{Literal}} & \textbf{\underline{Literal}} & \textbf{\underline{Literal}} & \textbf{Organ.} & \textbf{\underline{Literal}} \\
\textbf{Lebanese} (Language) & \textbf{Assoc.} & \textbf{\underline{Literal}} & \textbf{Demon.} & \textbf{Demon.} & \textbf{$--$} & \textbf{Misc.} & \textbf{$--$}\\
\textbf{Syrian} (Demonym) & \textbf{Assoc.} & \textbf{\underline{Literal}} & \textbf{Demon.} & \textbf{Demon.} & \textbf{$--$} & \textbf{Misc.} & \textbf{$--$}\\
\textbf{UK} origin (NounMod) & \textbf{Assoc.} & \textbf{Assoc.$^\dagger$} & \textbf{\underline{Literal}} & \textbf{\underline{Literal}} & \textbf{\underline{Literal}} & \textbf{\underline{Literal}} & \textbf{\underline{Literal}}\\
K.of \textbf{Jordan} (PostMod) & \textbf{Assoc.} & \textbf{Person} & \textbf{\underline{Literal}} & \textbf{\underline{Literal}} & \textbf{\underline{Literal}} & \textbf{Organ.} & \textbf{Person} \\
\textbf{Iraqi} militia (AdjMod) & \textbf{Assoc.} & \textbf{Assoc.$^\dagger$} & \textbf{Demon.} & \textbf{Demon.} & \textbf{$--$} & \textbf{Misc.} & \textbf{$--$}\\
\textbf{US} Congress (Embed) & \textbf{Assoc.} & \textbf{Organ.} & \textbf{Organ.} & \textbf{Organ.} & \textbf{Organ.} & \textbf{Organ.} & \textbf{Organ.} \\
\textbf{Turkey} (Metonymy) & \textbf{Assoc.} & \textbf{\underline{Literal}} & \textbf{\underline{Literal}} & \textbf{\underline{Literal}} & \textbf{\underline{Literal}} & \textbf{\underline{Literal}} & \textbf{$--$}\\
city in \textbf{Syria} (Literal) & \textbf{Literal} & \textbf{Literal} & \textbf{Literal} & \textbf{Literal} & \textbf{Literal} & \textbf{Literal} & \textbf{Literal}\\
\textbf{Iraqi} border (AdjMod) & \textbf{Literal} & \textbf{Literal} & \textbf{\underline{Demon.}} & \textbf{\underline{Demon.}} & \textbf{$--$} & \textbf{\underline{Misc.}} & \textbf{$--$}\\
\textbf{Min.of Defense} (Fac) & \textbf{Literal} & \textbf{\underline{Organ.}} & \textbf{\underline{Organ.}} & \textbf{\underline{Organ.}} & \textbf{\underline{Organ.}} & \textbf{\underline{Organ.}} & \textbf{\underline{Organ.}}\\\bottomrule
\end{tabular}
\caption{Popular NER taggers tested in June 2018 using official demo interfaces (incorrect labels \underline{underlined}) on the sentence: \textit{``\textbf{Milan}, who was speaking \textbf{Lebanese} with a \textbf{Syrian} of \textbf{UK} origin as well as the King of \textbf{Jordan}, reports that the \textbf{Iraqi} militia and the \textbf{US} Congress confirmed that \textbf{Turkey} has shelled a city in \textbf{Syria}, right on the \textbf{Iraqi} border near the \textbf{Ministry of Defense}."} A distinction is made only between \textit{a location} and \textit{not-a-location} since an \textit{associative label} is unavailable. The table shows only a weak agreement between tagging schemes. $^\dagger$Can be derived from the API with a simple rule.}
\label{comparetable}
\end{table*}

\paragraph{\bf Embedded Associative} toponyms are Non-Toponyms nested within larger entities such as \textit{``US} Supreme Court", \textit{``Sydney} Lottery" and \textit{``Los Angeles} Times". They are semantically, though not syntactically, equivalent to Associative Modifiers. Ignoring case would not change the meaning of the phrase \textit{``Nigerian} Army" versus \textit{``Nigerian} army". However, it \textit{will} wrongly change the shallow classification from ORG to LOC for most NER taggers.

\paragraph{\bf Homonyms}
and more specifically \textit{homographs}, are words with identical spelling but different meaning such as \textit{Iceland} (a UK grocery chain). Their meaning is determined mainly by contextual evidence \cite{hearst1991noun,gorfein2001activation} as is the case with other types. Examples include: ``\textit{Brooklyn} sat next to \textit{Paris}." and ``\textit{Madison, Chelsea, Clinton, Victoria, Jamison and Norbury} submitted a Springer paper." 

\section{Standard Evaluation Metrics}
\label{sef}
The previous section established \textit{what} is to be evaluated and \textit{why} it is important. In this part, we focus on critically reviewing existing geoparsing metrics, i.e. \textit{how} to assess geoparsing models. In order to reliably determine the SOTA and estimate the practical usefulness of these models in downstream applications, we propose a holistic, consistent and rigorous evaluation framework. Considering the task objective and available metrics, the recommended approach is to evaluate geoparsing as separate components. Researchers and practitioners do not typically tackle both stages at once \cite{delozier2015gazetteer,tobin2010evaluation,karimzadeh2013geotxt,wing2014hierarchical,wing2011simple,gritta2018melbourne}. More importantly, it is difficult to diagnose errors and target improvements without this separation. The best practice is to evaluate geotagging first, then obtain geocoding metrics for the true positives, i.e. the subset of correctly identified toponyms. We recommend evaluating with a minimum of 50\% of geotagged toponyms for a representative geocoding sample. Finally, population has not consistently featured in geocoding evaluation but it is capable of beating many existing systems \cite{delozier2015gazetteer,gritta2017s}. Therefore, we recommend the usage of this \textit{strong baseline} as a \textit{necessary component} of evaluation.

\subsection{Geotagging Metrics}
There is a strong agreement on the appropriate geotagging evaluation metric so most attention will focus on toponym resolution. As a subtask of NER, geotagging is evaluated using the \textit{F-Score}, which is also our recommended metric and an established standard for this stage of geoparsing \cite{lieberman2011multifaceted}. Figures for precision and recall may also be reported as some applications may trade precision for recall or may deem precision/recall errors more costly.

\subsection{Toponym Resolution Metrics}
Several geocoding metrics have been used in previous work and can be divided into \textit{three groups} depending on their output format. We assert that the most 'fit for purpose' output of a geoparser is a \textit{pair of coordinates}, not a categorical value or a ranked list of toponyms, which can give unduly flattering results \cite{santos2015using}. Ranked lists may be acceptable if subjected to further human judgement and/or correction but not as the final output. With set-based metrics such as the \textit{F-Score}, when used for geocoding, there are several issues: (a) Database incompatibility for geoparsers built with different knowledge bases that cannot be aligned to make fair benchmarking feasible. (b) The all-or-nothing approach implies that every incorrect answer (e.g. error greater than 5-10km) is equally wrong. This is not the case, geocoding errors are \textit{continuous} variables, not categorical variables hence the F-Score is unsuitable for toponym resolution. (c) Underspecification of recall versus precision, i.e. is a correctly geotagged toponym with an error greater than \textit{Xkm} a false positive or a false negative? This is important for accurate precision and recall figures. Set-based metrics and ranked lists are prototypical cases of trying to fit the wrong evaluation metric to a task. We now briefly discuss each metric group.

\paragraph{\bf Coordinates-based (continuous)} metrics are the recommended group when the output of a geoparser is a \textit{pair of coordinates}. An error is defined as the distance from predicted coordinates to gold coordinates. \textit{Mean Error} is a regularly used metric \cite{delozier2016data,hulden2015kernel}, analogous to a sum function thus informs of the total error as well. \textit{Accuracy@Xkm} is the percentage of errors resolved within \textit{Xkm} of gold coordinates. \cite{grover2010use} and \cite{tobin2010evaluation} used accuracy within 5km, \cite{santos2015using,dredze2013carmen} used accuracy at 5, 50, 250km, related works on tweet geolocation \cite{speriosu2013text,zheng2018survey,han2014improving,roller2012supervised} use accuracy at 161km. We recommend the more lenient 161km as it covers errors stemming from database misalignment. \textit{Median Error} is a simple metric to interpret \cite{wing2011simple,speriosu2013text} but is otherwise uninformative as the error distribution is non-normal hence not recommended. The Area Under the Curve \cite{gritta2017s,jurgens2015geolocation} is another coordinate-based metric, which follows in a separate subsection.

\paragraph{\bf Set-based/categorical} metrics and more specifically, the F-Score, has been used alongside coordinates-based metrics \cite{leidner2008toponym,andogah2010geographically} to evaluate the performance of the full pipeline. A true positive was judged as a correctly geotagged toponym \textit{and} one resolved to within a certain distance. This ranges from 5km \cite{andogah2010geographically,lieberman2012adaptive} to 10 miles \cite{kamalloo2018coherent,lieberman2010geotagging} to all of the previous thresholds \cite{kolkman2015cross} including 100km and 161km. In cases where WordNet has been used as the ground truth \cite{buscaldi2010toponym} an F-Score might be appropriate given WordNet's structure but it is not possible to make a comparison with a coordinates-based geoparser. Another problem with it is the all-or-nothing scoring. For example, \textit{Vancouver, Portland, Oregon} is an acceptable output if \textit{Vancouver, BC, Canada} was the expected answer. Similarly, the implicit suggestion that \textit{Vancouver, Portland} is equally wrong as \textit{Vancouver, Australia} is erroneous. Furthermore, using F-Score exclusively for the full pipeline does not allow for evaluation of individual geoparsing components making identifying problems more difficult. As a result, it is not a recommended metric for toponym resolution.

\paragraph{\bf Rankings-based} metrics such as Eccentricity, Cross-Entropy, Mean Reciprocal Rank, Mean Average Precision and other variants (Accuracy@k, Precision@k) have sometimes been used or suggested \cite{karimzadeh2016performance,craswell2009mean}. However, due to the aforementioned output format, ranked results are not recommended for geocoding. These metrics have erroneously been imported from Geographic Information Retrieval and should not be used in toponym resolution.

\begin{figure}[t]
    \centering
    \subfloat[Original Errors]{{\includegraphics[width=5.4cm]{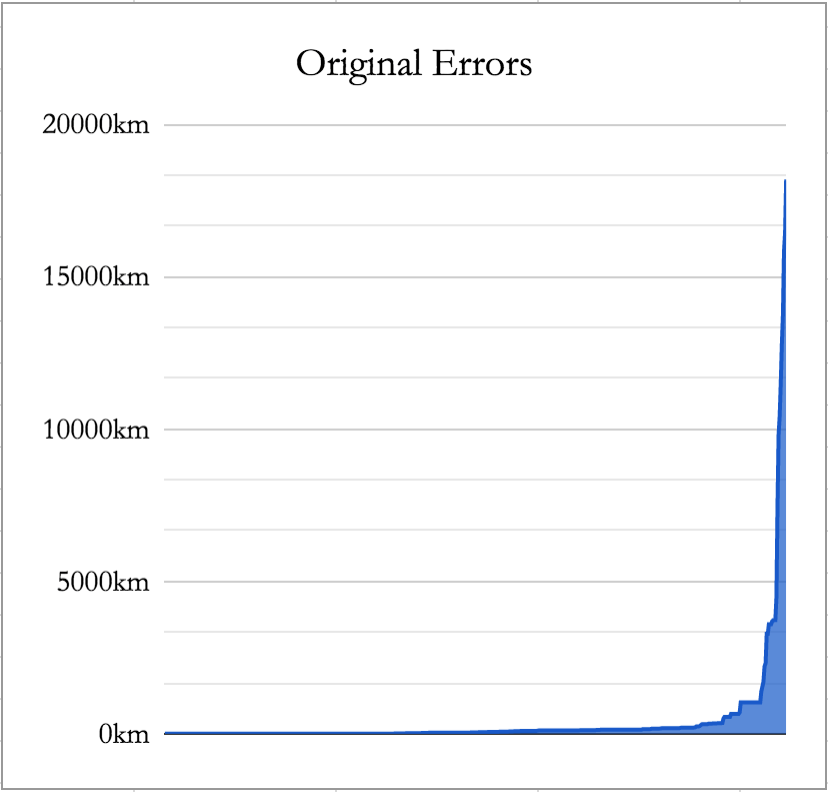} }}
    \qquad
    \subfloat[Logged Errors]{{\includegraphics[width=5.4cm]{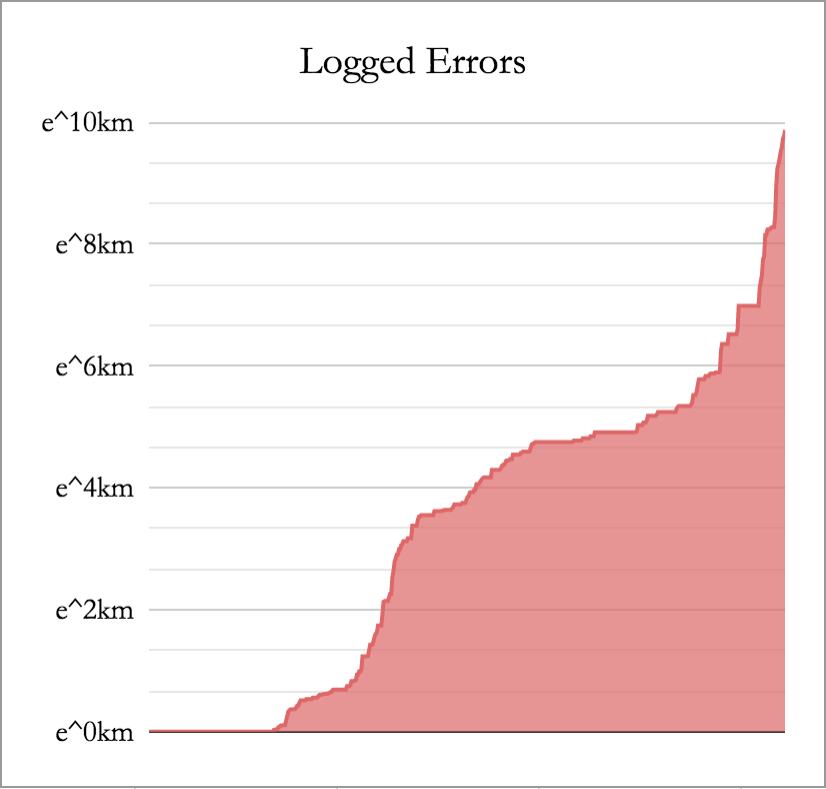} }}
    \caption{Computing the area under the curve by integrating the \textit{Logged Errors} in Figure (b). $AUC=0.33$ is interpreted as 33\% of the maximum geocoding error. 20,039 km is $1/2$ of Earth's circumference.}
    \label{auc}
\end{figure}

\paragraph{\bf Area Under the Curve (AUC)} is a recent metric used for toponym resolution evaluation \cite{gritta2017s,jurgens2015geolocation}. It is not to be confused with other AUC variants, which include the AUC of ROC, AUC for measuring blood plasma in Pharmacokinetics\footnote{The branch of pharmacology concerned with the movement of drugs within the body.} or the AUC of the Precision/Recall curve. The calculation uses the standard calculus method to integrate the area under the curve of geocoding errors denoted as \textit{x}, using the Trapezoid Rule\footnote{\url{https://docs.scipy.org/doc/numpy/reference/generated/numpy.trapz.html}}.
\begin{equation*}
\textit{Area ~Under ~the~ Curve} = \frac{\int_{0}^{dim(x)} \ln(x) dx}{dim(x) * ln(20039)}
\end{equation*}
 
The original errors, which are highly skewed in Figure \ref{auc}(a) are scaled down using the \textit{natural logarithm} resulting in Figure \ref{auc}(b). The area under the curve divides into the total area of the graph to compute the final metric value. The logarithm decreases the effect of \textit{outliers} that tend to distort the Mean Error. This allows for evaluation of the majority of errors that would otherwise be suppressed by outliers.

\subsection{Recommended Metrics for Toponym Resolution}
There is no single metric that covers every important aspect of geocoding, therefore based on the previous paragraphs, we make the following recommendations. (1) \textit{The AUC} is a comprehensive metric as it accounts for \textit{every error}, it is suitable for a rigorous comparison but needs some care to be taken to understand. (2) \textit{Accuracy@161km} is a fast and intuitive way to inform of ``correct" resolutions (error within 100 miles of gold coordinates) but ignores the rest of the error distribution. (3) \textit{Mean Error} is a measure of average and total error but it hides the full distribution, treats all errors as equal and is prone to distortion by outliers. Therefore, using \textit{all three metrics} gives a holistic view of geocoding performance as they compensate for each others' weaknesses while testing different aspects of toponym resolution. The SOTA model should perform well across all three metrics. As a final recommendation, an informative and intuitive way to assess the full pipeline would be to indicate how many toponyms were successfully extracted and resolved as in Table \ref{geocoding}. Using the Accuracy@161km, we can observe the \textit{percentage} of correctly recognised and resolved toponyms to estimate the performance of the combined system.

\subsection{Important Considerations for Evaluation}
\label{considerations}
\paragraph{The Choice of the Database} of geographic knowledge used by the geoparser and/or for labelling datasets must be clearly noted. In order to make a fair comparison between models \textit{and} datasets, the toponym coordinates must be a close match. Incompatibilities between global gazetteers have been previously studied \cite{acheson2017quantitative}. The most popular and open-source geoparsers and datasets do use Geonames\footnote{\url{https://www.geonames.org/export/}} allowing for an ``apples to apples" comparison (unless indicated otherwise).
In case it is required, we also propose a database alignment method for an empirically robust comparison of geoparsing models and datasets with incompatible coordinate data\footnote{The code can be found in the project's GitHub repository.}. The adaptation process involves a post-edit to the output coordinates. For each toponym, retrieve its nearest candidate by measuring the distance from the predicted coordinates (using a different knowledge base) to the Geonames toponym coordinates. Finally, output the Geonames coordinates to allow for a reliable comparison.

\paragraph{Resolution scope} also needs to be noted when comparing geoparsers, although it is less likely to be an issue in practice. Different systems can cover different areas, for example, geoparsers with \textit{Local Coverage} such as country-specific models \cite{matsuda2015annotating} versus \textit{Global Coverage}, which is the case with most geoparsers. It is not possible to fairly compare these two types of systems.

\paragraph{The train/dev/test data source domains,} i.e. the homogeneity or heterogeneity of the evaluation datasets is a vital consideration. The distribution of the evaluation datasets must be noted as performance will be higher on \textit{in-domain data}, which is when all partitions come from the same corpus. When training data comes from a different distribution from the test data, for example News Articles versus Wikipedia, the model that can \textit{generalise to out-of-domain test data} should be recognised as superior even if the scores are similar.

\paragraph{Statistical significance}\label{statistics} tests need to be conducted when making a comparison between two geoparsers unless a large performance gap makes this unnecessary. There are two options (1) \textit{k-fold cross-validation followed by a t-test} for \textit{both} stages or (2) the \textit{McNemar's test} for Geotagging and the \textit{Wilcoxon Signed-Rank Test} for Geocoding. The k-fold cross-validation is only suitable when a model is to be \textit{trained from scratch} on \textit{k-1} folds, \textit{k} times. For evaluation of trained geoparsers, we recommend using the latter options with similar statistical power, e.g. when it is infeasible to train several deep learning models.\\

K-Fold Cross-Validation works by generating 5-10 folds that satisfy the i.i.d. requirement for a parametric test \cite{dror2018hitchhiker}. This means folds should (a) come from disjoint files/articles and \textit{not} be randomised to satisfy the independent requirement and (b) come from \textit{the same domain} such as news text to satisfy the identically distributed requirement. GeoWebNews satisfies those requirements by design. The number of folds will depend on the size of the dataset, i.e. fewer folds for a smaller dataset and vice versa. Following that, we obtain scores for each fold, perform a t-test and report the p-value. There is a debate as to whether a p-value of 0.05 is rigorous enough. We think 0.01 would be preferred but in any case, the lower the more robust. Off-the-shelf geoparsers should be tested as follows.\\

For \textit{Geotagging}, use \textit{McNemar's test}, a non-parametric statistical hypothesis test suitable for matched pairs produced by binary classification or sequence tagging algorithms \cite{dietterich1998approximate}. McNemar's test compares the disagreement rate between two models using a contingency table of the outputs of two models. It computes the probability of two models 'making mistakes' at the same rate, using chi-squared distribution with one degree of freedom. If the probability of obtaining the computed statistic is less than 0.05, we reject the null hypothesis. For a more robust result, a lower threshold is preferred. This test is not well-approximated for contingency table values less than 25, however, if using multiple of our recommended datasets, this is highly unlikely.\\

For \textit{Toponym Resolution,} use a two-tailed \textit{Wilcoxon Signed-Rank Test} \cite{wilcoxon1945individual} for computational efficiency as the number of test samples across multiple datasets can be large (10,000+). Geocoding errors follow a power law distribution (Figure \ref{auc}a) with many outliers among the largest errors hence the non-parametric test. This sampling-free test compares the matched samples of geocoding errors. The null hypothesis assumes that the ranked differences between models' errors are centred around zero, i.e. model one is right approximately as much as model two. Finally, report the p-value and z-statistic. \\

\subsection{Unsuitable Datasets} 
Previous works in geoparsing \cite{leidner2004towards,andogah2010geographically,santos2015using,leidner2008toponym} have evaluated with their own labelled data but we have been unable to locate those resources. For those that are freely available, we briefly discuss the reasons for their unsuitability. AIDA \cite{hoffart2011robust} is a geo-annotated CoNLL 2003 NER dataset, however, the proprietary CoNLL 2003 data is required to build it. Moreover, the CoNLL file format does not allow for original text reconstruction due to the missing whitespace. SpatialML \cite{mani2010spatialml,mani2008spatialml} datasets are primarily focused on spatial expressions in natural language documents and are not freely available (\$500-\$1,000 for a license\footnote{\url{https://catalog.ldc.upenn.edu/LDC2011T02}}). Twitter datasets such as GeoCorpora \cite{wallgrun2018geocorpora} experience a gradual decline in completeness as users delete their tweets and deactivate profiles. WoTR \cite{delozier2016creating} and CLDW \cite{rayson2017deeply} are suitable only for digital humanities due to their historical nature and localised coverage, which is problematic to resolve \cite{butler2017alts}. CLUST \cite{lieberman2011multifaceted} is a corpus of clustered streaming news of global events, similar to LGL. However, it contains only 223 toponym annotations. TUD-Loc2013 \cite{katz2013learn} provides incomplete coverage, i.e. no adjectival or embedded toponyms, however, it may generate extra training data with some editing effort.

\subsection{Recommended Datasets} \label{recData}
We recommend evaluation with the following \textit{open-source} datasets: (1) WikToR \cite{gritta2017s} is a large collection of programmatically annotated Wikipedia articles and although quite artificial, to our best knowledge, it's the most difficult test for handling \textit{toponym ambiguity} (Wikipedia coordinates). (2) Local Global Lexicon (LGL) \cite{lieberman2010geotagging} is a global collection of local news articles (Geonames coordinates) and likely the most frequently cited geoparsing dataset. (3) GeoVirus \cite{gritta2018melbourne} is a WikiNews-based geoparsing dataset centred around disease reporting (Wikipedia coordinates) with global coverage though without adjectival toponym coverage. (4) TR-NEWS \cite{kamalloo2018coherent} is a new geoparsing news corpus of local and global articles (Geonames coordinates) with excellent toponym coverage and metadata. (5) Naturally, we also recommend GeoWebNews for a complete, fine-grained, expertly annotated and broadly sourced evaluation dataset.

\section{GeoWebNews}
\label{geowebnews}
As our final contribution, we introduce a new dataset to enable evaluation of fine-grained tagging and classification of toponyms. This will facilitate an immediate implementation of the proposals from previous sections. The dataset comprises 200 articles from 200 globally distributed news sites. Articles were sourced via a collaboration with the European Union's Joint Research Centre\footnote{\url{https://ec.europa.eu/jrc/en}}, collected during 1st-8th April 2018 from the European Media Monitor \cite{steinberger2013introduction} using a wide range of multilingual trigger words/topics\footnote{\url{http://emm.newsbrief.eu/}}. We then randomly selected exactly one article from each domain (English language only) until we reached 200 news stories. We also share the BRAT \cite{stenetorp2012brat} configuration files to expedite future data annotation using the new scheme. GeoWebNews can be used to evaluate the performance of NER (locations only) known as Geotagging and Geocoding/Toponym Resolution \cite{gritta2018melbourne}, develop and evaluate Machine Learning models for sequence tagging and classification, geographic information retrieval, even used in a Semantic Evaluation \cite{marquez2007semeval} task. GeoWebNews is a web-scraped corpus hence a few articles may contain duplicate paragraphs or some missing words from improperly parsed web links, which is typical of what might be encountered in practical applications.

\begin{figure}[t]
\centering
\includegraphics[width=\textwidth]{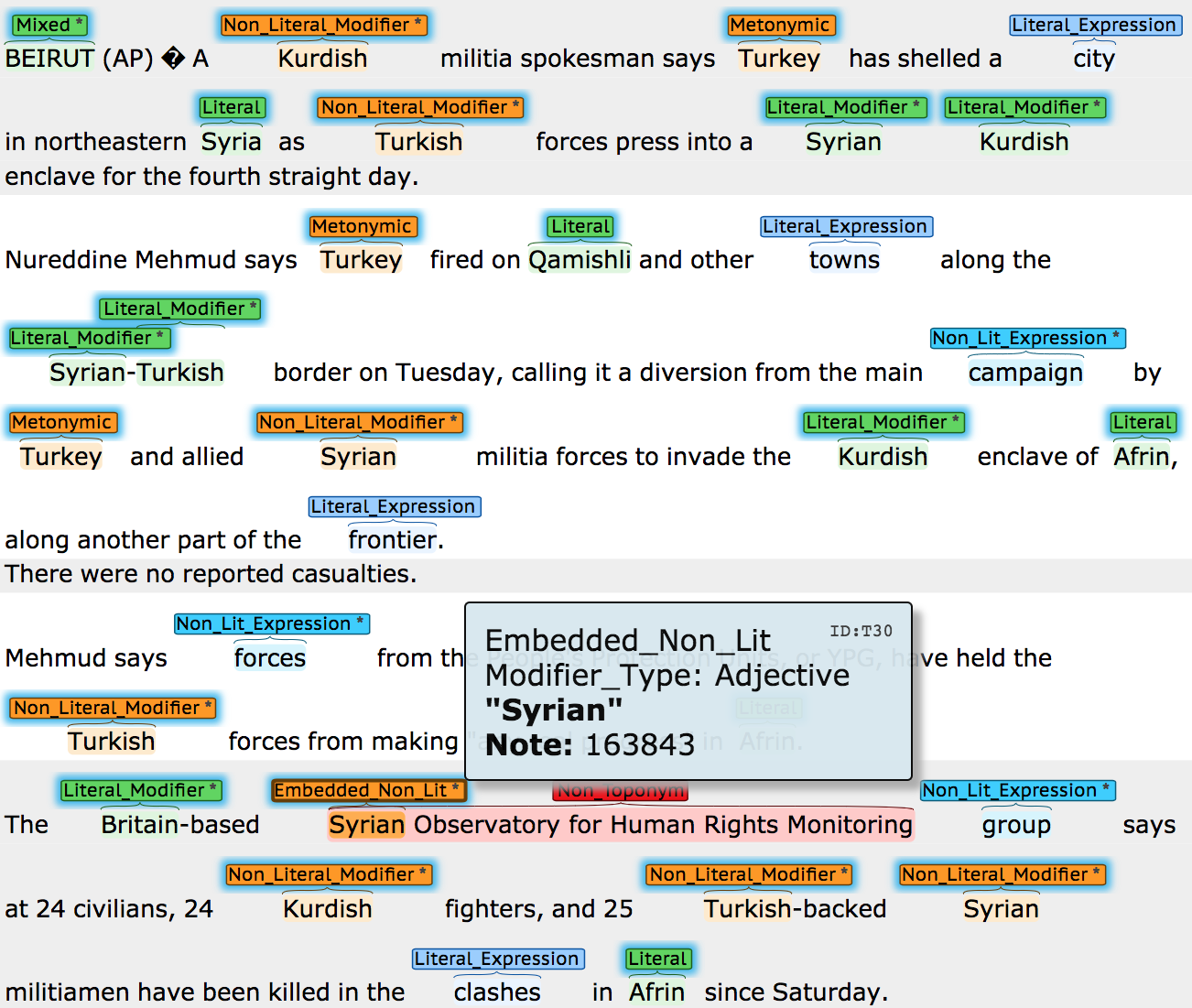}
\caption{A GeoWebNews article. An asterisk indicates an attribute, either \textit{a modifier\_type} [Adjective, Noun] and/or \textit{a non\_locational} [True, False].}
\label{annotation}
\end{figure}


\subsection{Annotation Procedure and Inter-Annotator Agreement (IAA)}
The annotation of 200 news articles at this level of granularity is a laborious and time-consuming effort. However, annotation quality is paramount when proposing changes/extensions to existing schemes. Therefore, instead of using crowd-sourcing, annotation was performed by the first author and two linguists from Cambridge University's Modern and Medieval Languages Faculty\footnote{\url{https://www.mml.cam.ac.uk/}}. An annotated article sample can be viewed in Figure \ref{annotation}. In order to expedite the verification process, we decided to make the annotations of the first author available to our linguists as `pre-annotation'. Their task was then twofold: (1) \textit{Precision Check}: verification of the first author's annotations with appropriate edits; (2) \textit{Recall Check}: identification of additional annotations that may have been missed. The F-Scores for the Geotagging IAA were computed using BratUtils\footnote{\url{https://github.com/savkov/BratUtils}}, which implements the MUC-7 scoring scheme \cite{chinchor1998appendix}. The Geotagging IAA after adjudication were 97.2 and 96.4 (F-Score), for first and second annotators respectively, computed on a 12.5\% sample of 336 toponyms from 10 randomly chosen articles (out of a total of 2,720 toponyms across 200 articles). The IAA for a simpler binary distinction (literal versus associative types) were 97.2 and 97.3.

\subsection{Annotation of Coordinates}
The Geocoding IAA with the first annotator on the same 12.5\% sample of toponyms expressed as accuracy [correct/incorrect coordinates] was 99.7\%. An additional challenge with this dataset is that some toponyms ($\sim$8\%) require either an extra source of knowledge such as Google Maps API, a self-compiled list of businesses and organisations names such as \cite{matsuda2015annotating} or even human-like inference to resolve correctly. These toponyms are facilities, buildings, street names, park names, festivals, universities and other venues. We have estimated the coordinates for these toponyms, which do not have an entry in Geonames using Google Maps API. These toponyms can be excluded from evaluation, which is what we did, due to the geoparsing difficulty. We have excluded 209 of these toponyms plus a further 110 demonyms, homonyms and language types without coordinates, evaluating with the remaining 2,401. We did not annotate the articles' geographic focus as was done for Twitter \cite{eisenstein2010latent,roller2012supervised} and Wikipedia \cite{laere2014georeferencing}. 

\subsection{Evaluation}
Sections \ref{taxonomy} and \ref{sef} have established GeoWebNews as a new standard dataset for fine-grained geoparsing grounded in real-world pragmatic usage. In the remainder of this section, we shall evaluate the SOTA Geoparsing and NER models to assess their performance on the linguistically nuanced location dataset, which should aid future comparisons with new NLP models. For a broad comparison, we have also included the Yahoo! Placemaker\footnote{The service was officially decommissioned but some APIs remain accessible.}, the Edinburgh Geoparser \cite{grover2010use} and our own CamCoder \cite{gritta2018melbourne} resolver as the main geoparsing benchmarks. We have also considered GeoTxt \cite{karimzadeh2013geotxt}, however, due to low performance, it was not included in the tables. Further related geoparsing evaluation with diverse datasets/systems can be found in our previous papers \cite{gritta2017vancouver}\cite{gritta2017s}.

\subsubsection{Geotagging GeoWebNews}
\label{geotaggingGWN}
For toponym extraction, we selected the two best models from Table \ref{comparetable}, Google Cloud Natural Language\footnote{\url{https://cloud.google.com/natural-language/}} and SpacyNLP\footnote{\url{https://spacy.io/usage/linguistic-features}}. We then trained an NCRF++ model \cite{yang2018ncrf++}, which is an open-source Neural Sequence Labeling Toolkit\footnote{\url{https://github.com/jiesutd/NCRFpp}}. We evaluated models using 5-Fold Cross-Validation (40 articles per fold, 4 train and 1 test fold). Embeddings were initialised with 300D vectors\footnote{Common Crawl 42B - \url{https://nlp.stanford.edu/projects/glove/}} from GloVe \cite{pennington2014glove} in a simple form of transfer learning as training data was limited. The NCRF++ tagger was trained with default hyper-parameters but with two additional features, the dependency head and the word shape, both extracted with SpacyNLP. For this custom model, we prioritised fast prototyping and deployment over meticulous feature/hyper-parameter tuning hence there is likely more performance to be found using this approach. The results are shown in Table \ref{emmGeotagging}.

\begin{table}[h]
\centering
\begin{tabular}{l|c|c|c}
\toprule
\textbf{NER Model / Geoparser}& \textbf{Precision} & \textbf{Recall} & \textbf{F-score} \\ \midrule
NCRF++ (Literal \& Associative labels) & 79.9 & 75.4 & 77.6 \\ \midrule
Yahoo! Placemaker & 73.4 & 55.5 & 63.2 \\
Edinburgh Geoparser & 81 & 52.4 & 63.6 \\
SpacyNLP & 82.4 & 68.6 & 74.9 \\
Google Cloud Natural Language & \textbf{91.0} & 76.6 & 83.2 \\
NCRF++ (``Location" label only) & 90.0 & \textbf{87.2} & \textbf{88.6} \\ \bottomrule
\end{tabular}
\caption{Geotagging F-Scores for GeoWebNews featuring the best performing models. The NCRF++ models' scores were averaged over 5 folds ($\sigma$=1.2-1.3).}
\label{emmGeotagging}
\end{table}

There were significant differences in precision and recall between off-the-shelf and custom models. SpacyNLP and Google NLP achieved a precision of 82.4 and 91 respectively while achieving a lower recall of  68.6 and 76.6 respectively. The NCRF++ tagger exhibited a balanced classification behaviour (90 precision, 87.2 recall). It achieved the highest F-Score of 88.6 despite only a modest amount of training examples.

\paragraph{\bf Geotagging with two labels} (physical location versus associative relationship) was evaluated with a custom NCRF++ model. The mean F-Score over 5 folds was 77.6 ($\sigma$=1.7), which is higher than SpacyNLP (74.9) with a single label. This demonstrates the feasibility of geotagging on two levels, treating toponyms separately in downstream tasks. For example, literal toponyms may be given a higher weighting for the purposes of geolocating an event. In order to incorporate this functionality into NER, training a custom sequence tagger is currently the best option for a two-label toponym extraction.

\subsubsection{Geocoding GeoWebNews}
\label{geocodingGWN}
For the evaluation of toponym resolution, we have excluded the following examples from the dataset. (a) the most difficult to resolve toponyms such as street names, building names, festival venues and so on, which account for $\sim$8\% of the total, without an entry in Geonames and often requiring a reference to additional resources. (b) demonyms, languages and homonyms, accounting for $\sim$4\% of toponyms as these are not locations hence do not have coordinates. The final count was 2,401 ($\sim$88\%) toponyms in the test set. Several setups were evaluated for a broad indication of expected performance. For geotagging, we used SpacyNLP to extract a \textit{realistic} subset of toponyms for geocoding, then scored the true positives with a matching entry in Geonames. The second geotagging method was Oracle NER, which \textit{assumes} perfect NER capability. Although artificial, it allows for geocoding of all 2,401 toponyms. We have combined these NER methods with the CamCoder \cite{gritta2018melbourne} default model\footnote{\url{https://github.com/milangritta/Geocoding-with-Map-Vector}}. The population heuristic was also evaluated as it was shown to be a strong baseline in our previous work. In practice, one should expect to lose up to 30-50\% toponyms during geotagging, depending on the dataset and NER. This may be seen as a disadvantage, however, in our previous work as well as in Table \ref{geocoding}, we found that a $\sim$50\% sample is representative of the full dataset.\\

\begin{table}[t]
\centering
\begin{tabular}{l|c|c|c|c}
\toprule
Setup/Description & Mean Err & Acc@161km & AUC & \# of Toponyms \\\midrule
SpacyNLP + CamCoder & \textbf{188} & \textbf{95} & \textbf{0.06} & 1,547 \\
SpacyNLP + Population & 210 & \textbf{95} & 0.07 & 1,547\\
Oracle NER + CamCoder & 232 & 94 & \textbf{0.06} & 2,401 \\
Oracle NER + Population & 250 & 94 & 0.07 & 2,401 \\
Yahoo! Placemaker* & 203 & 91 & 0.09 & 1444 \\
Edinburgh Geoparser* & 338 & 91 & 0.08 & 1363 \\ \bottomrule
\end{tabular}
\caption{Toponym Resolution scores for the GeoWebNews data. \textit{*This geoparser provides both Geotagging and Geocoding steps.}}
\label{geocoding}
\end{table}

The overall errors are low indicating low toponym ambiguity, i.e. low geocoding difficulty of the dataset. Other datasets \cite{gritta2017s} can be more challenging with errors 2-5 times greater. When provided with a database name for each extracted toponym (Oracle NER), it is possible to evaluate the whole dataset and get a sense of the pure disambiguation performance. However, in reality, geotagging is performed first, which reduces that number significantly. Using the geoparsing pipeline of SpacyNLP + CamCoder, we can see that 94-95\% of the 1,547 correctly recognised toponyms were resolved to within 161km. The number of recognised toponyms could be increased with a ``normalisation lexicon" that maps non-standard surface forms such as adjectives (``Asian", ``Russian", ``Congolese") to their canonical/database names. SpacyNLP provides a separate class for these toponyms called \textit{NORP}, which stands for nationalities, religious or political groups. Such lexicon could be assembled with a gazetteer-based statistical n-gram model such as \cite{al2017location} that uses multiple knowledge bases or a rule-based system \cite{volz2007towards}. For unknown toponyms, approximating the geographic representation from places that co-occur with it in other documents \cite{henrich2008determining} may be an option. Finally, not all errors can be evaluated in a conventional setup. Suppose an NER tagger has 80\% precision. This means 20\% of false positives will be used in downstream processing. In practice, this subset carries some unknown penalty that NLP practitioners hope is not too large. For downstream tasks, however, this is something that should be considered during error analysis.

\begin{figure}[h]
\centering
\includegraphics[width=0.99\textwidth]{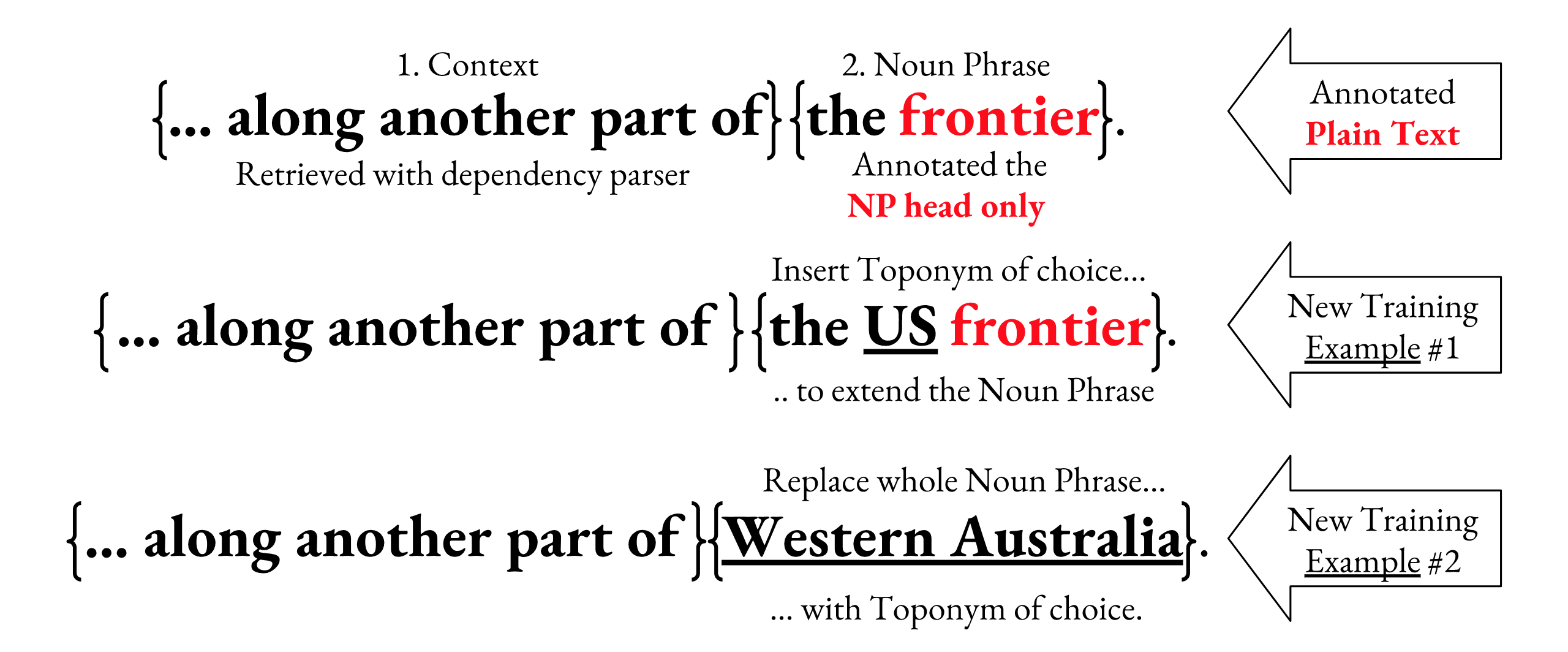}
\caption{An augmentation of a literal training example. An associative augmentation equivalent might be something like \textbf{\{The deal was agreed by\} \{the chief {\color{red} engineer}.\}} replacing "the chief engineer" by a toponym.}
\label{augmentation}
\end{figure}

\subsubsection{Training Data Augmentation}
\label{train}
We have built the option of data augmentation right into GeoWebnews and shall now demonstrate its possible usage in a short experiment. In order to augment the 2,720 toponyms to double or triple the training data size, two additional lexical features (\textit{NP heads}) were annotated, denoted \textit{Literal Expressions} and \textit{Associative Expressions}\footnote{Google Cloud NLP already tags common nouns in a similar manner.}. These annotations generate two separate components (a) the NP context and (b) the NP head itself. In terms of distribution, we have literal (N=1,423) versus associative (N=2,037) \textit{context} and literal (N=1,697) versus associative (N=1,763) \textit{heads}, indicated by a binary \textit{non-locational} attribute. These two interchangeable components give us multiple permutations from which to generate a larger training dataset\footnote{\url{https://github.com/milangritta/Pragmatic-Guide-to-Geoparsing-Evaluation}} (see Figure \ref{augmentation} for an example). The associative expressions are deliberately dominated by ORG-like types because this is the most frequent metonymic pair \cite{alonso2013annotation}.

\begin{table}[t]
\centering
\begin{tabular}{c|c|c|c}
\toprule
(1) No Aug. & (2) Partial Aug. & (3) Full Aug. & (4) Ensemble of (1, 2, 3) \\ \midrule
\textbf{88.6} & 88.2 & 88.4 & 88.5 \\ \bottomrule
\end{tabular}
\caption{F-Scores for NCRF++ models with 5-Fold Cross-Validation. No improvement was observed for the augmented or ensemble setups over baseline.}
\label{ncrfpp}
\end{table}

Table \ref{ncrfpp} shows three augmentation experiments (numbered 2, 3, 4) that we have compared to the best NCRF++ model (1). We hypothesised that data augmentation, i.e. adding additional modified training instances would lead to a boost in performance, however, this did not materialise. An ensemble of models (4) also did not beat the baseline NCRF++ model (1). Due to time constraints, we have not extensively experimented with elaborate data augmentation and \textit{encourage further research} into other implementations.

\section{Conclusions}

\subsection{Future Work}
Geoparsing is a special case of NER and often the initial step of an information extraction pipeline used by downstream applications. A detailed use case of the benefits of geoparsing and our pragmatic taxonomy of toponyms can be seen in Chapter 6 (page 95) of this PhD thesis \cite{gritta2019you}. Geoparsing is a key step for the monitoring of the spread of public health threats such as epidemics and food-borne diseases using public news text. The chapter shows how fine-grained toponym extraction enables a deeper understanding and classification of geographic events using deep learning models with SOTA performance, significantly improving upon previous approaches. This methodology lets researchers automatically learn about entities associated with particular geographic areas. The ideas proposed in our paper can therefore enable a more accurate analysis of geographic events described in free text. Whether it is public health risks or other domains of interest, in the age of Big Data, there is a need for automated information processing of relevant events at scale.\\

We also expect to see more (1) reproduction/replication studies to test and/or revise evaluation setups, (2) dataset/model probing to test the validity of SOTA results, and (3) the annotation of multilingual and multi-domain resources for a wider range of tasks. Examples include a recently published paper on Twitter \textit{user geolocation} \cite{mourad2019practical} where the authors provide a critical review of current metrics, systems, datasets and SOTA claims. Similar to our work, the authors also recommend the use of several metrics for a holistic evaluation of user geolocation. Another consideration that applies beyond geoparsing is the construction of standard dataset splits for evaluation, investigated in 'We need to talk about standard splits' \cite{gorman-bedrick-2019-need}. The authors reproduced several SOTA part-of-speech models evaluated on standard 80-10-10 train-dev-test splits. However, when the splits were \textit{randomly} generated, the SOTA rankings were not reliably reproduced. With that in mind, our practical guide to geoparsing evaluation complies with this recommendation as the cross-validation was performed with 5 folds generated from randomly sampled news articles. \\

It is critical that automatic evaluation is closely aligned with human evaluation and that this is periodically examined as we have done in this paper. Incorrectly structured datasets can also produce misleading comparisons with human performance. In 'Probing Neural Network Comprehension of Natural Language Arguments' \cite{niven2019probing}, the authors carefully examined BERT's \cite{devlin2018bert} peak performance on the argument reasoning task. The 77\% accuracy was \textit{only 3 points} below an untrained human baseline. However, it transpired that this performance came from exploiting the dataset's patterns, rather than the model's language understanding ability. The authors then created an adversarial version by removing those regularities resulting in just 53\% accuracy, slightly above random choice. It is therefore prudent to ensure the robustness of evaluation and caution against any premature claims of near-human or superhuman performance\footnote{\url{https://gluebenchmark.com/leaderboard/}}. \\

In section \ref{geowebnews}, we introduced a new dataset for fine-grained geoparsing. However, we also encourage future efforts to be focused on corrections to existing datasets (with the consultation of expert linguists, if possible) such as CoNLL 2003 \cite{tjong2003introduction}. Many models still benchmark their performance on the original (non-random) splits \cite{yadav2018survey}, for example at COLING \cite{yang2018design} and ACL \cite{gregoric2018named}. A survey/review could keep the original 3-class annotation, utilise our taxonomy to make the dataset suitable for geoparsing evaluation or even extend the taxonomy to other NER classes. An example of a dataset correction is MultiWOZ 2.1 \cite{eric2019multiwoz}, which is frequently used for training and evaluation of dialogue systems. The authors made changes to over 32\% of state annotations across 40\% of dialogue turns, which is a significant correction to the original dataset \cite{budzianowski2018multiwoz}. The final future work proposal is a Semantic Evaluation task in the Information Extraction track to close the gap to human (expert) baselines, almost 100\% for geocoding (95\% for SOTA) and around 97 F-Score for geotagging (87 for SOTA). GeoWebNews is most suitable for sequence labelling evaluation of the latest machine learning models. It comes with an added constraint of limited training samples, which could be overcome with transfer learning via pretrained language models such as BERT or ELMo \cite{peters2018deep}.

\subsection{Closing Thoughts}
The Principle of Wittgenstein's Ruler from Nassim N. Taleb's book, Fooled by Randomness \cite{taleb2005fooled} deserves a mention as we reflect on the previous paragraphs. It says: \textit{``Unless you have confidence in the ruler's reliability, if you use a ruler to measure a table you may also be using the table to measure the ruler.}" In the field of NLP and beyond, this translates into: ``\textit{Unless you have confidence in the reliability of the evaluation, if you use the tools (data, metrics, splits, etc.) to evaluate models, you may also be using the models to evaluate the tools.}" We must pay close attention to the representativeness of the evaluation methods. It is important to ask whether models 'successfully' evaluated with tools that do not \textit{closely mirror} real-world conditions and human judgement is the goal to aim for in NLP.\\

In this manuscript, we introduced a detailed pragmatic taxonomy of toponyms as a way to increase Geoparsing recall and to differentiate literal uses (53\%) of place names from associative uses (47\%) in a corpus of multi-source global news data. This helps clarify the task objective, quantifies type occurrences, informs of common NER mistakes and enables innovative handling of toponyms in downstream tasks. In order to expedite future research, address the lack of resources and contribute towards replicability and extendability \cite{goodman2016does,cacho2018reproducible}, we shared the annotation framework, recommended datasets and any tools/code required for fast and easy extension. The NCRF++ model trained with just over 2,000 examples showed that it can outperform SOTA taggers such as SpacyNLP and Google NLP for location extraction. The NCRF++ model can also achieve an F-Score of 77.6 in a two-label setting (literal, associative) showing that fine-grained toponym extraction is feasible. Finally, we critically reviewed current practices in geoparsing evaluation and presented our best recommendations for a holistic and intuitive performance assessment. As we conclude this section, here are the recommended evaluation steps.

\begin{enumerate}
\item Review (and report) important geoparsing considerations in Section \ref{considerations}.
\item Use a proprietary or custom NER tagger to extract toponyms using the recommended dataset(s) as demonstrated in Section \ref{geotaggingGWN}.
\item Evaluate geotagging using F-Score as the recommended metric and report statistical significance with McNemar's Test (Section \ref{statistics}). 
\item Evaluate toponym resolution using Accuracy@161, AUC and Mean Error as the recommended metrics, see Section \ref{geocodingGWN} for an example.
\item \textit{Optional}: Evaluate geocoding in ``laboratory setting" as per Section \ref{geocodingGWN}.
\item Report the number of toponyms resolved and the statistical significance using the Wilcoxon Signed-Rank Test (Section \ref{statistics}).
\end{enumerate}

\bibliographystyle{spmpsci} 
\bibliography{template}
\end{document}